%% file: ms.tex
\pdfoutput=1

\documentclass{article}

\input{pre.tex}

\usepackage{spconf,amsmath,graphicx}

\title{Convex relaxations of convolutional neural nets}
\name{Burak Bartan\quad\quad Mert Pilanci}
\address{Department of Electrical Engineering\\ Stanford University}
\begin{document}
%
\maketitle
\begin{abstract}
\input{abstract.tex}
\end{abstract}
\begin{keywords}
Convolutional neural networks, convex relaxations, linear programming, deep learning
\end{keywords}

\section{Introduction}\label{sec:intro}
\input{intro}

\subsection{Related Work and Contributions}
\input{related_work}

\section{Relaxations for a Single Neuron}
\input{single_neuron}

\vspace{-0.4cm}
\section{Convolutional nets and multiple neurons}
\input{multiple_neurons}

\vspace{-0.5cm}
\section{Numerical Results}
\vspace{-0.3cm}
\input{numerical_results}

\vspace{-0.4cm}
\section{Conclusion}
\input{conclusion}

\vfill\pagebreak

\bibliographystyle{IEEEbib}
\bibliography{ms.bbl}

\end{document}

%% file: pre.tex
\usepackage[dvipsnames]{xcolor}
\usepackage{everysel}
\usepackage{tikz}
\usepackage{pgfplots}
\usepackage{epstopdf}
\usepackage{graphicx}
\usetikzlibrary{fit}
\pgfplotsset{compat=newest} 
\pgfplotsset{plot coordinates/math parser=false} 
\usetikzlibrary{shapes,arrows}
\usetikzlibrary{shapes.geometric}
\usetikzlibrary{calc}

 \tikzset{%
  blocks/.style    = {draw, thick, rectangle, minimum height = 3em,
    minimum width = 3em},
  sum/.style      = {draw, circle, node distance = 2cm}, 
  input/.style    = {coordinate}, 
  output/.style   = {coordinate} 
  buffer/.style={
        draw,
        shape border rotate=180,
        regular polygon,
        regular polygon sides=3,
        fill=white,
        node distance=2cm,
        minimum height=4em
    }

}
\newif\ifNightMode
\NightModetrue
\NightModefalse
\ifNightMode
\EverySelectfont{\color{CornflowerBlue}}
\everymath{\color{LimeGreen}}
\color[rgb]{1,1,1}
\color{LimeGreen}
\else
\fi
\usepackage{fancyhdr}
\usepackage{amsmath}
\usepackage{dsfont}
\usepackage{amssymb}
\usepackage{color}
\usepackage{graphicx}
\usepackage{amsfonts}
\usepackage{float}
\usepackage{url,graphicx,tabularx,array,geometry,amsmath,fullpage,appendix}
\usepackage{enumerate}

\usepackage{amsmath, amsthm, amssymb}
\newtheorem{theorem}{Theorem}

\newcommand{\what}{\hat w}
\newcommand{\bcar}{\begin{carlisthack}}
\newcommand{\ecar}{\end{carlisthack}}

\newcommand{\real}{\mathbb{R}}

\newcommand{\eye}{I}
\newcommand{\N}{\mathcal{N}}
\newcommand{\wstar}{w^*}

\DeclareMathOperator{\Cone}{cone}
\DeclareMathOperator{\cone}{cone}

\usepackage[symbol]{footmisc}

\usepackage[dvips,all]{xy}

\long\def\comment#1{}

%% file: abstract.tex
We propose convex relaxations for convolutional neural nets with one hidden layer where the output weights are fixed. For convex activation functions such as rectified linear units, the relaxations are convex second order cone programs which can be solved very efficiently. We prove that the relaxation recovers the global minimum under a planted model assumption, given sufficiently many training samples from a Gaussian distribution. We also identify a phase transition phenomenon in recovering the global minimum for the relaxation. 

%% file: intro.tex
Convolutional neural networks (CNNs) have been extremely successful across many domains in machine learning and computer vision \cite{krizhevsky2012imagenet,lecun2015deep}. However, analyzing the behavior of non-convex optimization methods used to train CNNs remains a challenge. In this paper, we propose finite dimensional convex relaxations for convolutional neural nets which are composed of a single hidden layer with convex activation functions. We prove that for the ReLU activation function, a randomized perturbation of the convex relaxation recovers the global optimum with probability approaching $\frac{1}{2}$, which can be amplified to $1-\big(\frac{1}{2}\big)^r$ using $r$ independent trials of random perturbations. We illustrate a phase transition phenomenon in the probability of recovering the global optimum. Finally, we consider  learning filters for a regression task on the MNIST dataset of handwritten digits \cite{lecun1998gradient}.



%% file: related_work.tex
In recent years, there has been an increasing amount of literature on providing theoretical results for neural networks. A considerable amount of work in this area focused on the case where a convolutional network with a single hidden layer is trained using gradient descent. For instance, \cite{brutzkus2017convnet} shows that gradient descent achieves the global minimum (with high probability) for convolutional networks with one layer and no overlap when the distribution of the input is Gaussian. \cite{brutzkus2017convnet} also proves that the problem of learning this network is NP-complete. Zhang et. al propose a convex optimization approach based on a low-rank relaxation using the nuclear norm regularizer \cite{zhang2016convexified}. In \cite{askari2018lifted}, Askari et al. consider neural net objectives which are convex over blocks of variables. A number of recent results considered the gradient descent method on the non-convex training objective, and proved that it recovers the planted model parameters under distributional assumptions on the training data \cite{du2017gradient,soltanolkotabi2017learning,zhang2018learning}. We refer the reader to \cite{goel2016reliably,goel2018learning,du2018many} for other theoretical results regarding learning ReLU units and convolutional nets.

Our contributions can be summarized as follows. First, we propose a randomized convex relaxation of learning single hidden layer neural networks in the original parameter space. This should be contrasted with \cite{zhang2016convexified}, where the convex program is in the lifted space of matrices, and is not guaranteed to find the global optimum. Our derivation also explains why direct relaxations fail and a randomized perturbation is needed. Second, we prove that the relaxation obtains the global optimum under a planted model assumption with Gaussian training data with certain probability. Our results are geometric in nature, and has a close connection to the phase transitions in compressed sensing \cite{CandesTao05,Donoho06,Donoho13,donoho2009observed,rudelson2008sparse}, escape from a mesh phenomena in Gaussian random matrices \cite{gordon1988milman,amelunxen2013living} and sketching \cite{PilWai14a,PilWai14b,pilanci2017newton,yang2017randomized}. Third, our numerical results highlight a phase transition, where the global optimum can be recovered when the number of samples exceeds a threshold that depends on the dimension of the filter and the number of hidden neurons. Our approach provides a general framework for obtaining convex relaxations which can be used in other architectures such as soft-max classifiers, autoencoders and recurrent nets.










%% file: single_neuron.tex
Consider the problem of fitting a single neuron to predict a continuous labels $y$ from observations $x$. Suppose that we observe the training data $\{ x_i \}_{i=1}^n$ and labels $\{ y_i \}_{i=1}^n$.
\begin{align}
p^* = \min_{w \in \real^d} \, \sum_{i=1}^n (\sigma( x_i^T w) - y_i)^2\,, \label{eqn:oneNeuron}
\end{align}
where $\sigma(\cdot)$ is the activation function. In general, the optimization of this objective for an arbitrary training set is computationally intractable. We refer the reader to recent works on the NP-hardness of ReLU regression and approximation algorithms \cite{DeyRelu}.
We rewrite the objective in \eqref{eqn:oneNeuron} by introducing an additional slack variable as follows
\begin{align}
p^* = &\min_{w \in \real^d,\, z \in \real^n} \, \sum_{i=1}^n (z_i - y_i)^2\,, \\
& z_i = \sigma( x_i^T w)\,,i=1,...,n
\label{eqn:oneNeuronAux}
\end{align}
and consequently we relax the equality constraint into an inequality constraint and obtain
\begin{align}
p^* \ge p =  &\min_{w, z} \, \sum_{i=1}^n (z_i - y_i)^2\,. \\
& z_i \ge \sigma( x_i^T w)\,,i=1,...,n
\label{eqn:oneNeuronRelax}
\end{align}
The above problem is a finite dimensional convex optimization problem which can be solved very efficiently \cite{Boyd02}.
\subsection{Failure of the naive relaxation}
As a result of the relaxation, the convex optimization problem in \eqref{eqn:oneNeuronRelax} may not be a satisfactory approximation of the original problem, and may have more than one optimal solution. Let us illustrate the case for the ReLU activation $\sigma(u) = (u)_+$. The convex relaxation \eqref{eqn:oneNeuronRelax} becomes 
\begin{align}
 p_{ReLU} = &\min_{w, z} \, \sum_{i=1}^n (z_i - y_i)^2\,  \\
& z_i \ge  (x_i^T w)_+\,,i=1,...,n \nonumber
\label{eqn:oneNeuronRelaxRelu}\\
=&\min_{w, z} \, \sum_{i=1}^n (z_i - y_i)^2\,, \\
& z_i \ge  x_i^T w\,,i=1,...,n \nonumber\\
& z_i \ge 0\,,i=1,...,n\, \nonumber
\end{align}
where we have used the fact that $(u)_+\le z$ holds if and only if $u \le z$ and $0 \le z$. Observe that $w=0$ and $z=y$ is feasible in the constraint set of \eqref{eqn:oneNeuronRelaxRelu}, and also minimizes the convex objective $\|y-z\|_2^2$. Hence, the pair $w=0$, $z=y$ belongs to the set of optimal solutions of \eqref{eqn:oneNeuronRelaxRelu}, regardless of the data $x_1,...,x_n$ and labels $y_1,...,y_n$. Let us define $\wstar$ as the solution to \eqref{eqn:oneNeuron}, and suppose that the optimal value is zero. Note that the pair $z= y= (X\wstar)_+$ and $w=\wstar$ also belongs to the set of optimal solutions. Therefore, the set of optimal solutions to the convex program \eqref{eqn:oneNeuronRelaxRelu} is not a singleton, and always contains the trivial solution, along with the optimal solution to \eqref{eqn:oneNeuron}.

It is surprising that even in the idealized case where the labels are generated by $y=(X\wstar)_+$, and features are i.i.d. Gaussian distributed, i.e., $x_i \sim \N(0,\eye_d)$, $i=1,...,n$, the above relaxation fails to recover the correct weight vector $\wstar$. In contrast, it is possible to recover $\wstar$ using a very simple procedure as long as $n>2d$, and $n$ is large enough. We can consider the subset of labels which are strictly positive, $S=\bigcup_{i\in\{1,...,\} \,:\, y_i>0} \{i\}$,
%
%
and attempt to solve for $\wstar$ using the pseudoinverse via $\what = X_S^{\dagger} y_S$\,. It's straightforward to show that $\what = \wstar$ as long as $|S|\ge d$ and $X_S$ has full column rank. For i.i.d. Gaussian features, this holds with high probability when $n$ is large enough.

%
%

\subsection{Randomized Convex Relaxation}
In convex relaxations, relaxing the equality constraints into inequality constraints can lead to multiple spurious feasible points. It is clear that with the naive convex relaxation, we can't hope to recover the optimal solution to \eqref{eqn:oneNeuron}. In this section we will propose a randomly perturbed convex program aiming to recover the solution of the original problem. The reasoning behind optimizing a random objective is to randomly pick a solution among multiple feasible solutions. We will pick a random vector $r$ distributed as $\N(0,\eye_d)$ and solve 
%
%
\begin{align}
\what = \arg &\min_{w \in \real^d,\, z \in \real^n} \,\frac{1}{2}\|z-y\|_2^2 +  \beta \,r^T w\,, \label{eqn:oneNeuronRand} \\
& z_i \ge \sigma( x_i^T w)\,,i=1,...,n\,, \nonumber
\end{align}
where $\beta>0$ is a small regularization parameter that controls the amount of random perturbation. For the ReLU activation, the above is a second-order cone program which can be solved efficiently \cite{Boyd02}. The next theorem shows that a small random perturbation allows exact recovery of the global optimum as $\beta \rightarrow 0$.
\begin{theorem} \label{single_neuron_thm}
Let $\sigma(u)= \max(u,0)$, the ReLU activation, $x_i \sim N(0,\eye_d)$, $i=1,...,n$ and the global minimum value of \eqref{eqn:oneNeuron} is $p^*=0$, and is achieved\footnote[1]{In other words, the response $y$ is generated by a network such that $y=\sigma(x_i^T \wstar)$, $i=1,...,n$ holds. This is a common assumption which is also used by many others in the literature, and sometimes referred as the teacher network assumption.} by $w^*$. Then, the solution of \eqref{eqn:oneNeuronRand} as $\beta \rightarrow 0$ is equal to the global minimizer $\wstar$ with probability $\frac{1}{2}-c_2 e^{-c_3 d}$ if $n\ge c_1 d$, where $c_1,c_2,c_3$ are universal constants. 
\end{theorem}
\noindent This theorem essentially implies that a random perturbation to the objective is equally likely to return $w=w^*$ or $w=0$, which are the only extreme points asymptotically as $n\rightarrow \infty$.\\
{\it Proof sketch:}
Plugging in the ReLU activation function $\sigma(u) = (u)_+$ we can express the optimization problem as $\beta \rightarrow 0$ via a linear program
\begin{align*}
&\min \, r^T w\\
& s.t.~ y = z,\, Xw \le z\,, 0 \le z\,.
\end{align*}
The pair $z=y=(X\wstar)_+$ and $w=\wstar$ are optimal if
\begin{align*}
(y-z) + r= r \in \cone X_S\,.
\end{align*}
%
%
The dual linear program is given by 
\begin{align*}
&\min_{u\ge 0}\, y^T u\\
& X^Tu = r\,.
\end{align*}
\vspace{-0.3cm}
Optimality conditions for the primal-dual pair are as follows
\begin{align*}
\exists u \,:\, u\ge 0,\, u_{S^c}=0,\, X^Tu = r\,,
\end{align*}
where $S = \bigcup_{i\in\{1,...,n\} \,:\, x_i^T \wstar >0} \{i\}$ is the subset of indices over $\{1,...,n\}$ where the ReLU is active.
This condition can be equivalently represented as a cone intersection 
%
\begin{align*}
r \in \Cone(X^T_S)\,,
\end{align*}
where $X_S$ is the submatrix of $X$ composed of the rows that are in $S$. Without loss of generality, let us take $\wstar = e_1$ due to the rotation invariance of the i.i.d. Gaussian distribution over the features. In this case, we have $S=\{i\,:\,x_{i1}>0\}$ and $S$ is a random set independent of $x_{ij}$ for $j \neq 0$. Partitioning $r$ and $X_S$ into $r=[r_1;r^\prime]$ and $X_S^T = [x_{1S};X^\prime_S]$, the cone condition reduces to
$r_1 = x_{1S}^T u_S$, and $r^\prime =  (X^\prime_{S})^T u_S$\,.
Fixing $r_1>0$ and $x_{1S}$, the above condition can be stated as the nullspace-cone intersection $[u_S;1] \in {\mathrm{Null}}([X^\prime_S\, -r])$ and $[x_{1S}^T;-r_1]^T[u_S;1]=0$, $u_S\ge0$. Since $[X^\prime_S\, -r]$ is an i.i.d. Gaussian matrix, the nullspace-cone intersection probability can be lower-bounded by calculating the Gaussian width of the $\cone\big([u_S;1]: [x_{1S}^T;-r_1]^T[u_S;1]=0, u_S\ge 0  \big)$ \cite{amelunxen2013living}. For $r_1>0$, a calculation of the Gaussian width yields $O(d)$, which implies that for $n\ge c_1 d$, the probability of cone intersection is $1-c_2e^{-c_3 d}$ conditioned on $r_1\ge 0$. Noting that $P(r_1>0) = \frac{1}{2}$, we obtain the claimed result.

%% file: multiple_neurons.tex
We now consider multiple neurons where each neuron receives the output of a convolution.
%
When the filter $w$ is applied in a non-overlapping fashion, we need to solve the non-convex problem $\min_{w,c} \ \sum_{i=1}^n (y_i-\sum_{j=1}^k c_j (X_{ij}w)_+)^2$, where $X_{ij} \in \mathbb{R}^{1\times d/k}$ is the $j$'th $\frac{d}{k}$-length block of the $i$'th row of $X$. If we assume that $c \geq 0$, then without loss of generality, we can instead solve the problem $\min_{w} \ \sum_{i=1}^n (y_i-\sum_{j=1}^k (X_{ij}w)_+)^2$. This follows from the fact that $c_j (X_{ij}w)_+ = (c_j X_{ij} w)_+$ for $c_j \geq 0$ and therefore it is possible to implicitly include the parameter $c_j$ in the parameter $w$. 
We relax this non-convex problem the same way we did for the single neuron case and obtain the relaxed linear program as we let $\beta \rightarrow 0$
\begin{align} \label{relaxed_multiple}
&\max_{z_{ij}\geq 0, w} \, r^Tw\\
& \sum_{j=1}^k z_{ij} = y_i, i = 1,...,n \nonumber \\
& z_{ij} \geq X_{ij}w, \forall (i,j) \in \{1,...,n\} \times \{1,...,k\}. \nonumber
\end{align}
The dual of (\ref{relaxed_multiple}) is given by
\begin{align} \label{dual_multiple}
&\min_{\lambda_{ij} \geq 0, v} \, y^Tv\\
& \lambda_{ij} \leq v_i, \forall (i,j) \in \{1,...,n\} \times \{1,...,k\} \nonumber \\
& \sum_{i=1}^n \sum_{j=1}^k X_{ij}^T \lambda_{ij} = r, \nonumber
\end{align}
where $\lambda \in \mathbb{R}^{n \times k}$ and $v \in \mathbb{R}^n$. Defining the sets $S_j$ for $j = 1,...,k$ as $S_j = \bigcup_{i\in \{1,...,n\} \,:\, X_{ij} \wstar >0} \{i\}$, we can rewrite the equality constraint of (\ref{relaxed_multiple}) as
\begin{align*}
    \sum_{j=1}^k \sum_{i \in S_j} X_{ij}^T \lambda_{ij} + \sum_{j=1}^k \sum_{i \in S_j^c} X_{ij}^T \lambda_{ij} = r.
\end{align*}
Substituting $\lambda_{ij}$ with $v_i + (\lambda_{ij} - v_i)$ and multiplying both sides by $\wstar$, we obtain
\begin{align} \label{wstar_sum}
    \sum_{j=1}^k \sum_{i \in S_j} X_{ij} \wstar v_i  +  \sum_{j=1}^k \sum_{i \in S_j} X_{ij} \wstar (\lambda_{ij} - v_i) \nonumber \\ +  \sum_{j=1}^k \sum_{i \in S_j^c} X_{ij} \wstar \lambda_{ij} = r^T \wstar.
\end{align}
Note that the first term of the LHS of (\ref{wstar_sum}) is equal to the objective of the dual (this follows from the definition that $ReLU(X_{ij} \wstar) = 0$ for $i \not \in S_j$), and the RHS is equal to the optimal value of the primal. For optimality, we must have the second and the third terms of the LHS to be zero since they are both nonpositive. This implies that $\lambda_{ij} = 0$ for all $i \not \in S_j$, and $\lambda_{ij}=v_i$ for all $i \in S_j$.

Hence, the optimality condition is that there exists $v \in \mathbb{R}^n$ such that
\begin{align} \label{opt_cond}
    \sum_{j=1}^k X_{S_j, j}^T v_{S_j} = r, \text{ and } \ v_{S_j} \geq 0, \ j = 1,...,k,
\end{align}
where $v_{S_j}$ is a $|S_j|$-dimensional vector with entries from $v$ corresponding to the indices $S_j$. To reach an equivalent condition to (\ref{opt_cond}) that will be useful in our analysis, let us define the sets $R_i$ for each sample $i = 1,...,n$, which indicate the indices $j$ for which the $i$'th sample satisfies $X_{ij} \wstar > 0$:
\begin{align}
    R_i = \bigcup_{j\in \{1,...,k\} \,:\, i \in S_j} \{j\}
\end{align}
The condition (\ref{opt_cond}) is satisfied if and only if $r$ falls inside the cone defined by the vectors $ \sum_{j\in R_i}X_{ij}^T$, $i = 1,...,n$, that is,
\begin{align} \label{cone_cond}
    r \in \cone\left(\bigcup_{i\in \{1,...,n\} } \Bigg\{ \sum_{j\in R_i}X_{ij}^T  \Bigg\} \right).
\end{align}
Now, we are ready to show that (\ref{cone_cond}) is satisfied with probability $\frac{1}{2}-c_2e^{-c_3d}$ if $n\geq c_1d$, where $c_1,c_2$, and $c_3$ are constants (not necessarily the same constants for the single neuron case in Theorem 1). Assuming $X$ and $r$ both have i.i.d. $\mathcal{N}(0,1)$ entries, and observing that we can consider only the samples with $|R_i|=1$, we can use the same reasoning in Theorem \ref{single_neuron_thm}. The expected number of samples with $|R_i|=1$ is $n \frac{1}{2} (\frac{1}{2})^{k-1} = \frac{n}{2^k}$. For a fixed $k$, the number of samples with $|R_i|=1$ is linear in $n$, and the constant term $\frac{1}{2^k}$ may as well be hidden in the constant $c_1$. Now using the same argument from Theorem \ref{single_neuron_thm}, it is straightforward to show that the success probability is $\frac{1}{2}-c_2e^{-c_3d}$ for $n\geq c_1d$.

%% file: numerical_results.tex
In this section, we present the results of numerical experiments. The first subsection serves to visualize and compare the phase transition phenomena for our proposed relaxation method and gradient descent. In all of the experiments, the filter is applied with no overlaps, thus the filter size is equal to $d/k$. The second subsection presents numerical experiments on the MNIST dataset.
\vspace{-0.4cm}
\subsection{Phase Transition Plots for Gaussian Distribution}
\begin{figure}[htb]
\begin{minipage}[b]{0.48\linewidth}
  \centering
  \centerline{\includegraphics[width=3.7cm]{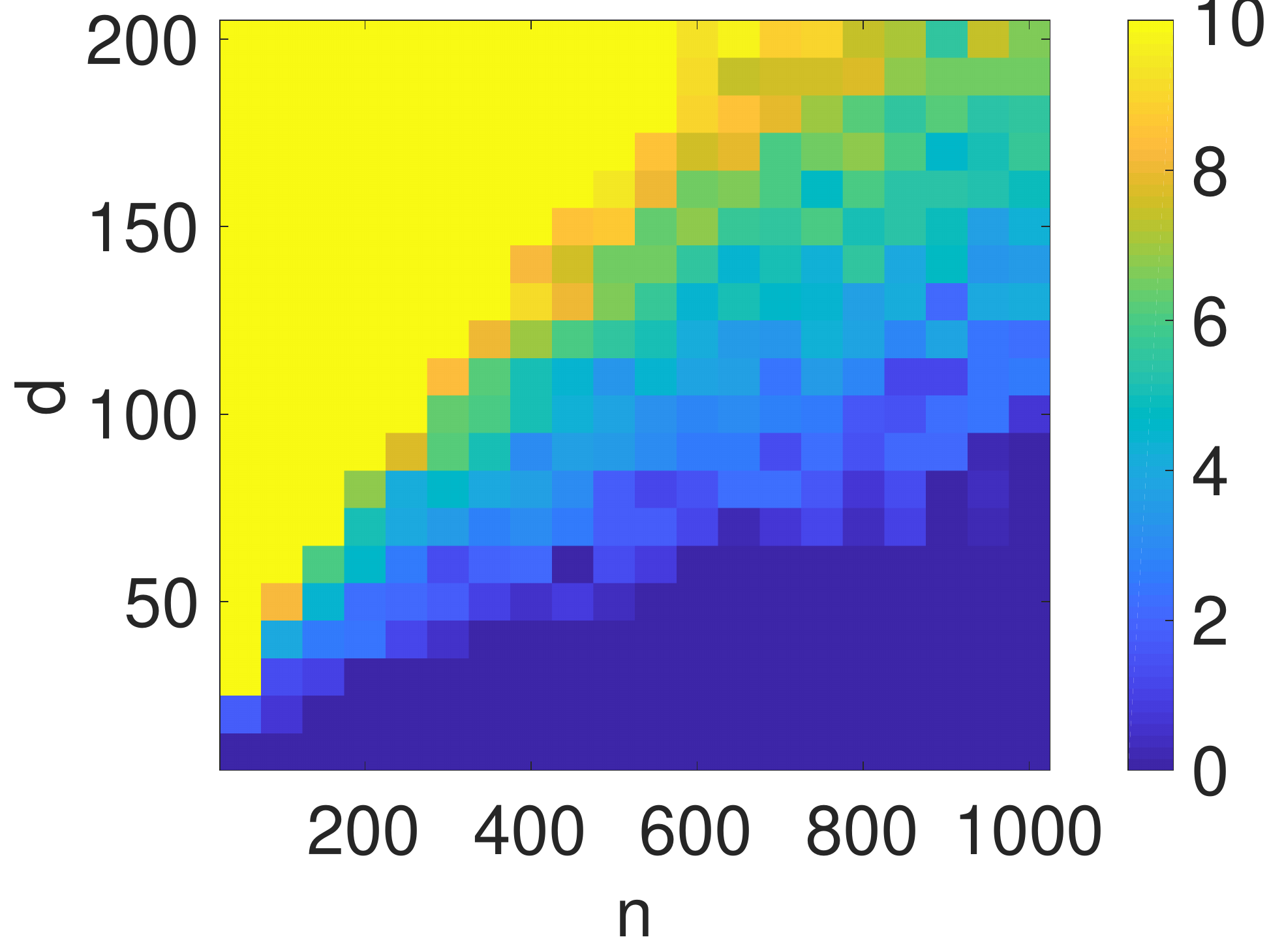}}
  \centerline{(a) CNN relaxation, k = 1}\medskip
\end{minipage}
\hfill
\begin{minipage}[b]{0.48\linewidth}
  \centering
  \centerline{\includegraphics[width=3.7cm]{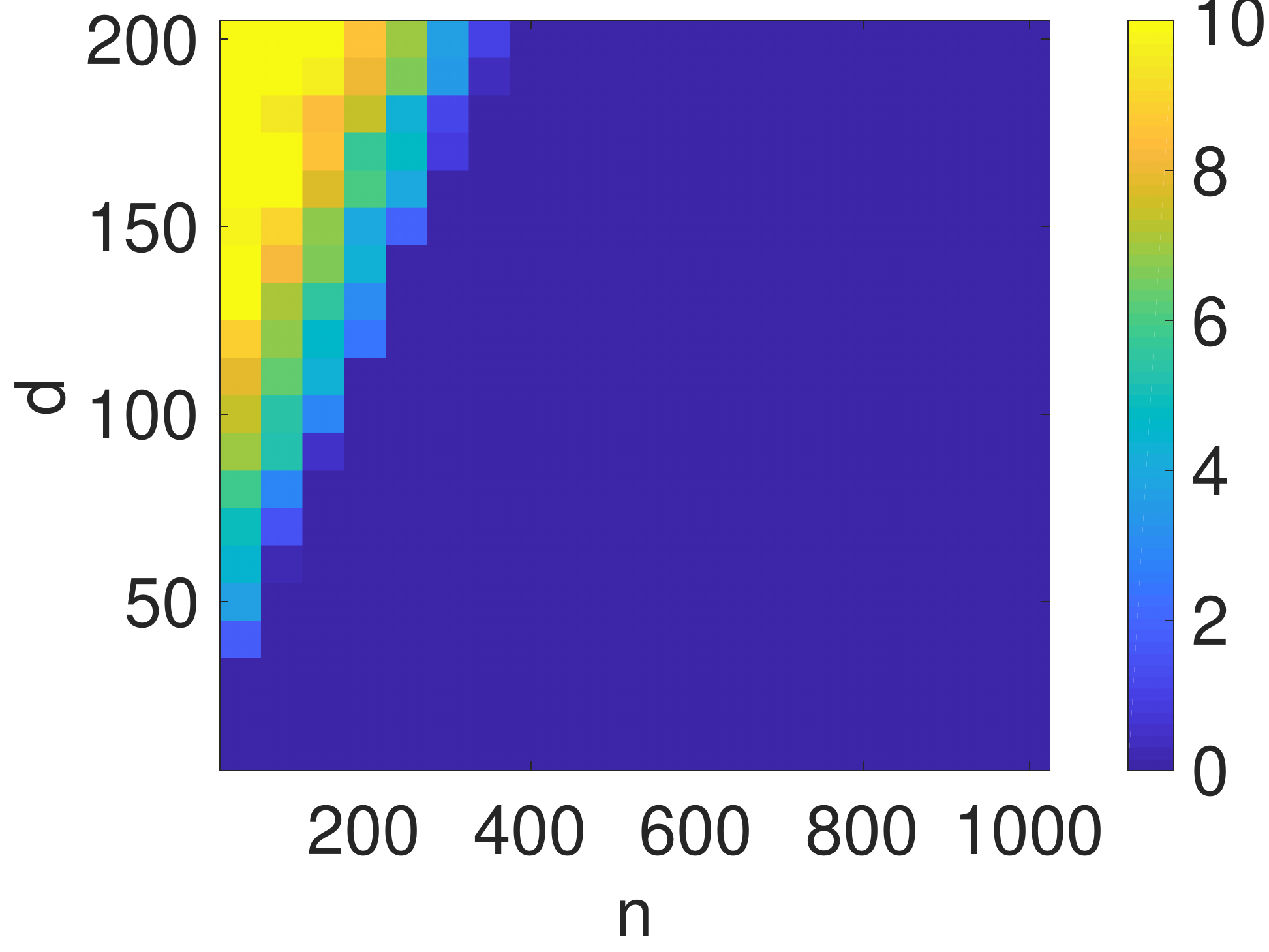}}
  \centerline{(b) Gradient descent, k = 1}\medskip
\end{minipage}
\hfill
\begin{minipage}[b]{.48\linewidth}
  \centering
  \centerline{\includegraphics[width=3.7cm]{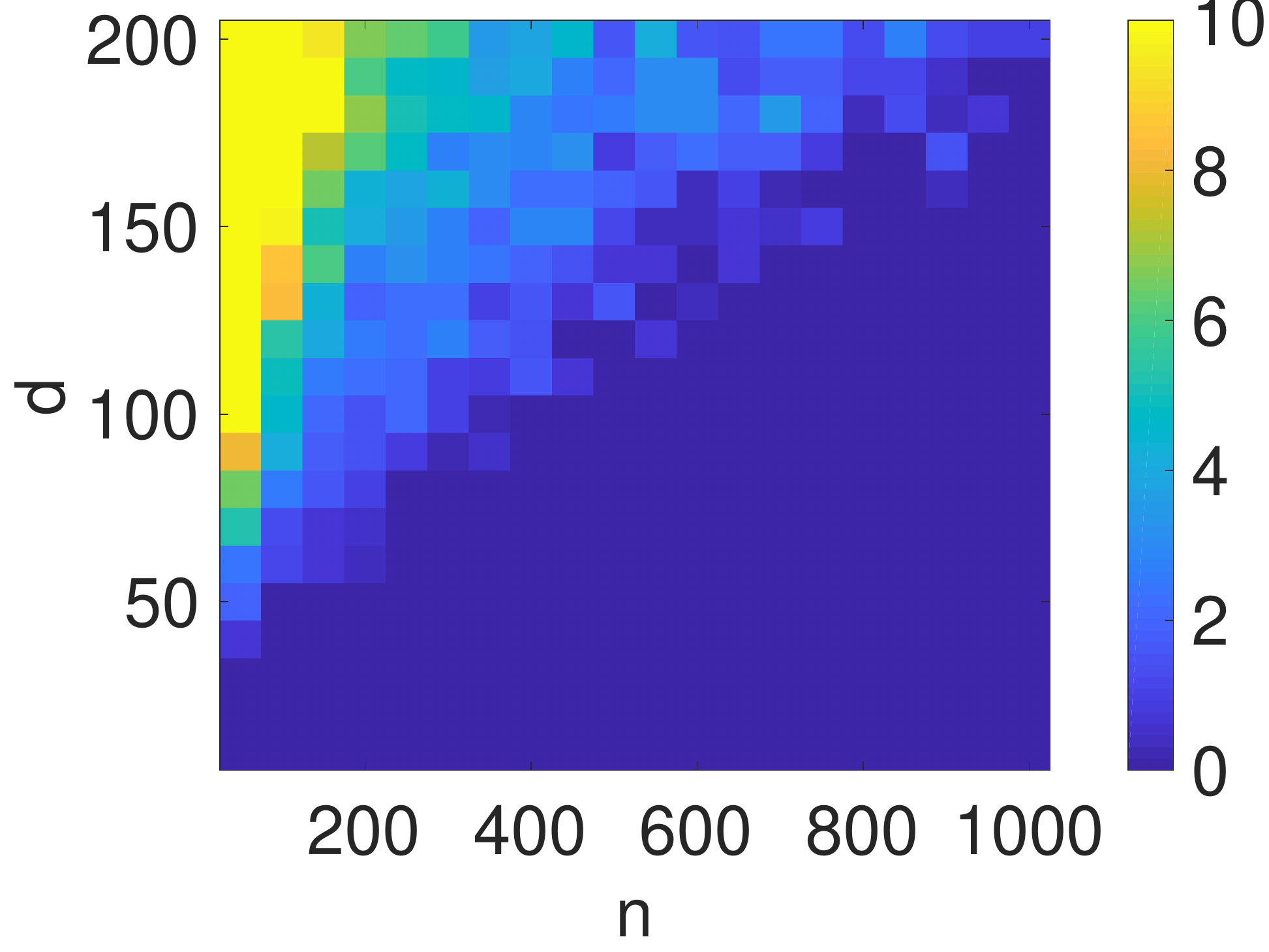}}
  \centerline{(c) CNN relaxation, k = 2}\medskip
\end{minipage}
\hfill
\begin{minipage}[b]{0.48\linewidth}
  \centering
  \centerline{\includegraphics[width=3.7cm]{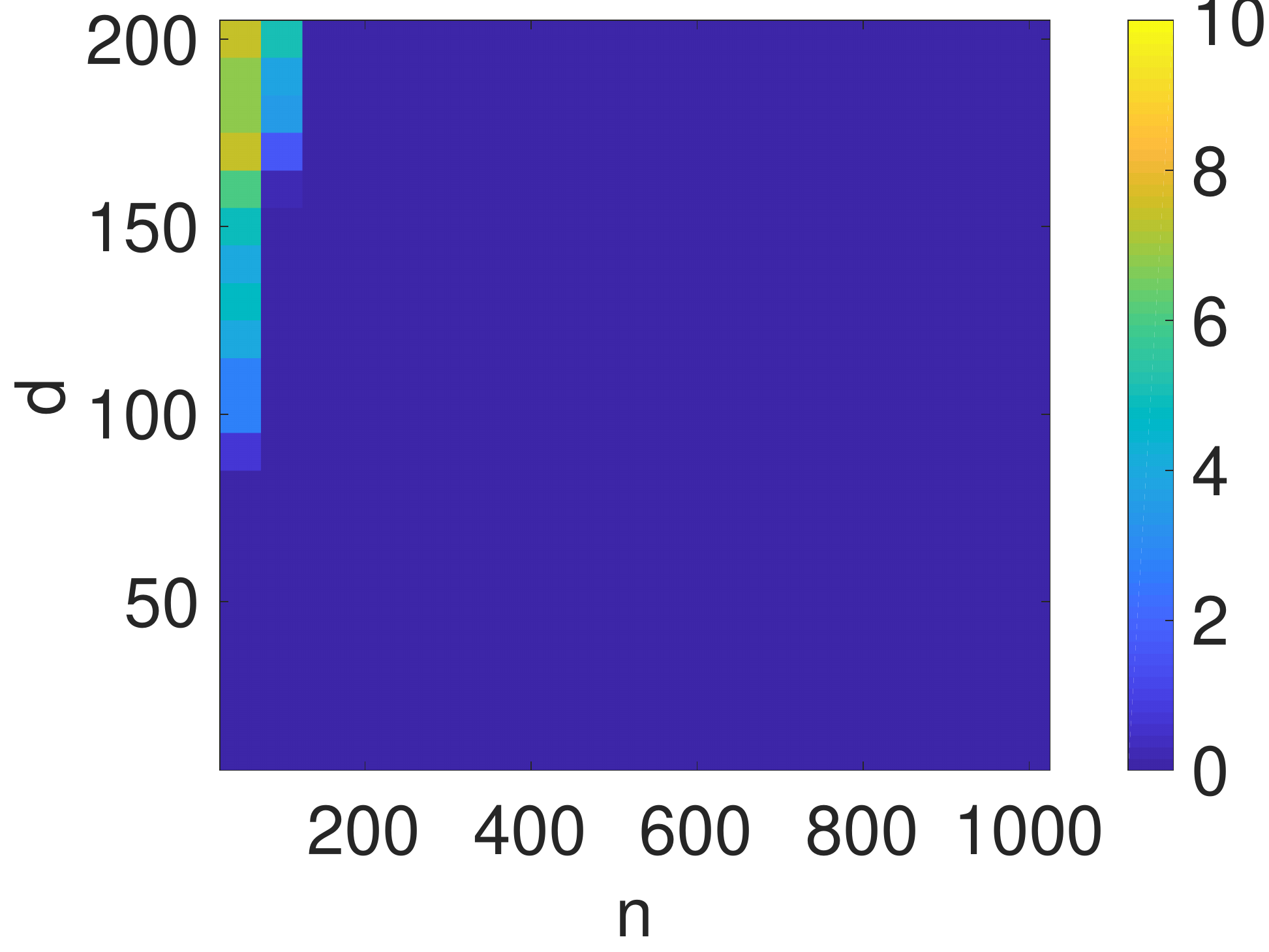}}
  \centerline{(d) Gradient descent, k = 2}\medskip
\end{minipage}
\hfill
\begin{minipage}[b]{0.48\linewidth}
  \centering
  \centerline{\includegraphics[width=3.7cm]{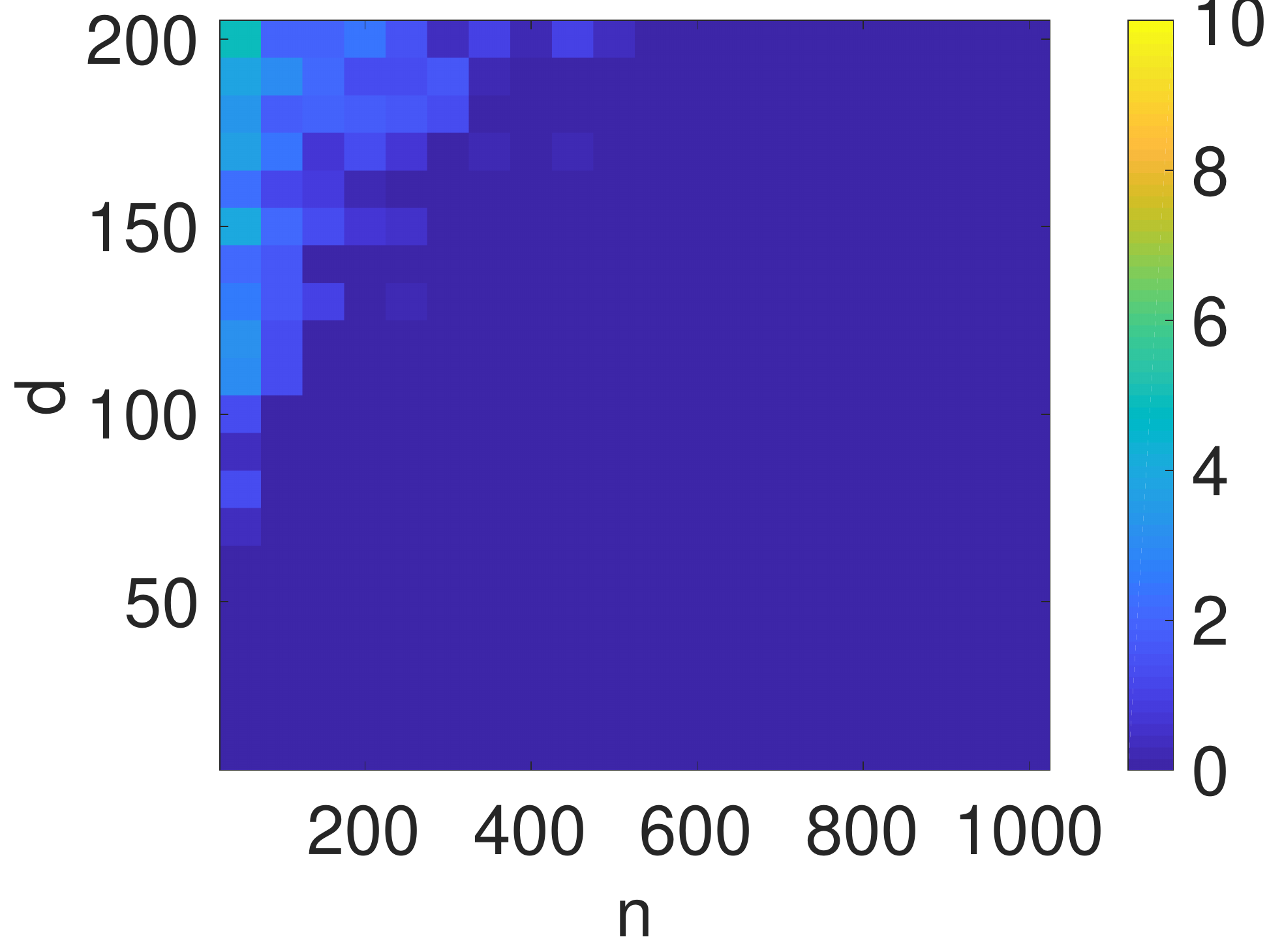}}
  \centerline{(e) CNN relaxation, k = 5}\medskip
\end{minipage}
\hfill
\begin{minipage}[b]{0.48\linewidth}
  \centering
  \centerline{\includegraphics[width=3.7cm]{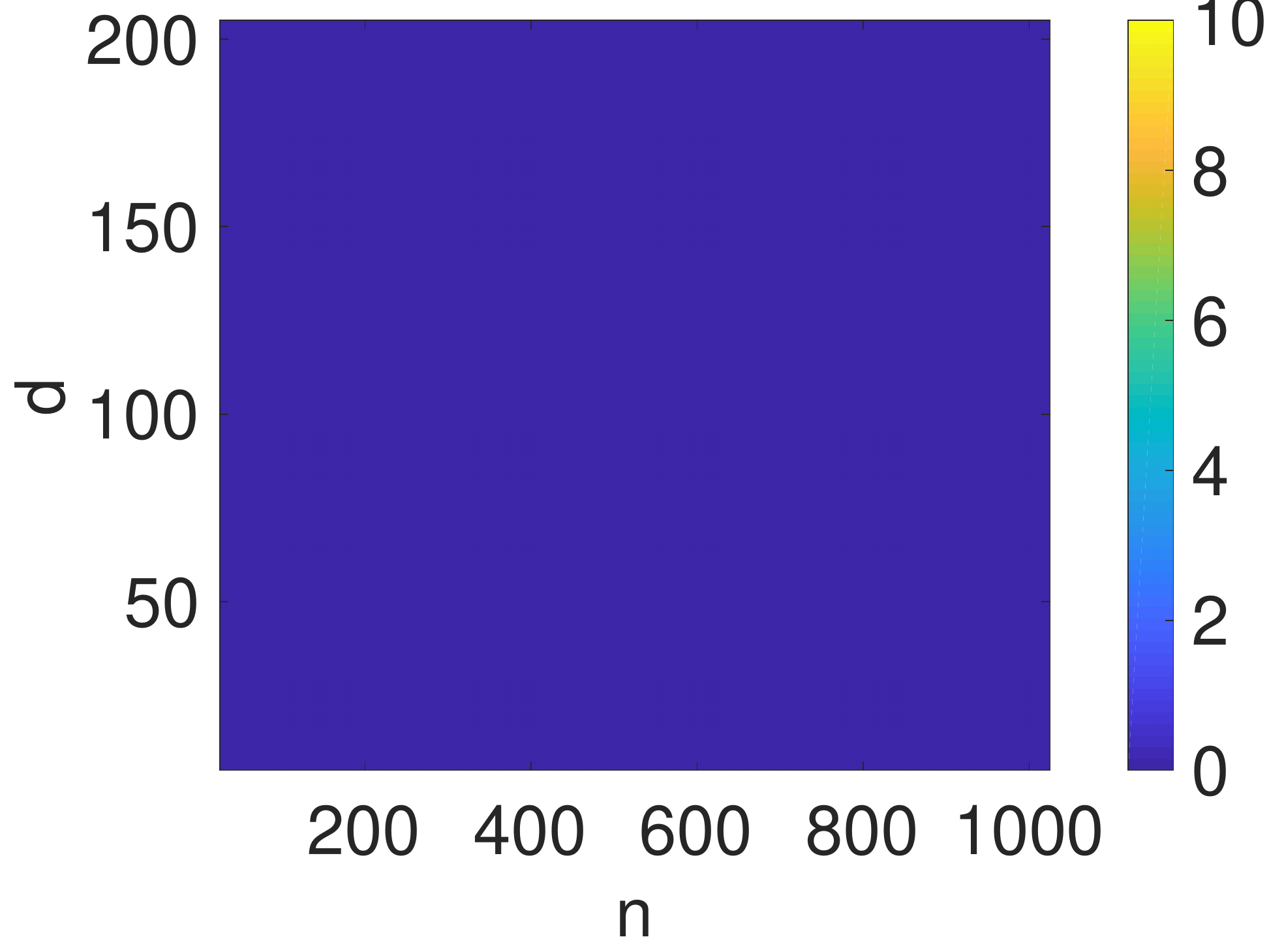}}
  \centerline{(f) Gradient descent, k = 5}\medskip
\end{minipage}
\vspace{-0.5cm}
\caption{Phase transition plots comparing CNN relaxation and gradient descent when the training data is i.i.d. Gaussian.}
\label{fig:phase_trans_gaussian}
\vspace{-0.4cm}
\end{figure}

For all the phase transition plots in Fig. \ref{fig:phase_trans_gaussian}, for given $n$ and $d$, we generate a random data matrix $X$, and a random filter $\wstar$ and compute the output $y$ without adding noise (i.e., teacher network assumption). Then we run the proposed relaxation method and gradient descent algorithms separately and repeat it 100 times for each method. The plots show the minimum of the resulting 100 $l_2$-norm errors between $\wstar$ and $\what$, that is $||\wstar - \what||_2$, using scaled colors. Fig. \ref{fig:phase_trans_gaussian} illustrates the performances of the proposed relaxation method and gradient descent for $k=1,2,5$ neurons when the distribution of the input is Gaussian, $\mathcal{N}(0,I_d)$. Fig. \ref{fig:phase_trans_gaussian} shows that as $k$ increases, the probabilities of recovering $\wstar$ go up for both methods. For a given $k$, gradient descent seems to outperform the proposed relaxation method as the line where the phase transition occurs has a higher slope. However, gradient descent does not offer the same flexibility the proposed method does since the proposed relaxation is a convex problem and can handle extra convex constraints. Fig. \ref{fig:phase_trans_gaussian} also shows that as $k$ increases, the difference between the performances of the convex relaxation and gradient descent vanishes. We believe that this implies that the non-convex loss surface of the convolutional nets is becoming more like a convex surface as $k$ increases.
\vspace{-0.3cm}

\vspace{-0.2cm}
\subsection{Experiments on MNIST Dataset}
We now present the experiment results on a randomly rotated version of the MNIST dataset \cite{lecun1998gradient}, where the task is to predict the rotation angles of handwritten digits. The results are given in Table \ref{tab:mnist_table}. We compare two methods where we perform least squares (LS) with $l_2$ regularization, on different features. For the baseline we use the original pixels, and in the second method we augment the original pixels with the filtered features where the filter is obtained by fitting the proposed relaxation to $y$.
\begin{table}[h]
\vspace{-0.5cm}
  \begin{center}
    \caption{MNIST experiment results.}
    \label{tab:mnist_table}
    \begin{tabular}{l|r} 
      \textbf{Experiment} & \textbf{RMSE} \\
      \hline
      LS using raw pixels& 17.04 \\
      LS with learned filter & 16.59 \\
    \end{tabular}
  \end{center}
  \vspace{-0.6cm}
\end{table}
The results in Table \ref{tab:mnist_table} show root-mean-square errors (RMSE) for the predicted rotation angles. These results have been obtained on the test set, which has 5000 images unseen to the training process. Table \ref{tab:mnist_table} shows that the proposed relaxation method helps extract useful features that make the model generalize better to the test data.

%% file: conclusion.tex
\vspace{-0.2cm}
We investigated convex relaxations for single hidden layer no-overlap convolutional neural nets. We proved that under the planted model assumption, the relaxation method finds the global optimum with probability $\frac{1}{2}$. It is possible to make the success probability arbitrarily close to 1 by running the algorithm multiple times. We gave phase transition plots to help identify the behavior of the proposed convex relaxation in different parameter regimes.  We also presented a numerical study on a real dataset and empirically showed that the proposed convex relaxation is able to extract useful features. We believe this work provides insights towards understanding the relationship between the non-convex nature of deep learning and convex relaxations.
